\documentclass[10pt,twocolumn,letterpaper]{article}

\usepackage{iccv}
\usepackage{times}
\usepackage{epsfig}
\usepackage{graphicx}
\usepackage{amsmath}
\usepackage{amssymb}

\usepackage[pagebackref=true,breaklinks=true,letterpaper=true,colorlinks,bookmarks=false]{hyperref}
\usepackage[ruled,linesnumbered]{algorithm2e}
\usepackage{multirow}
\usepackage{arydshln}

\usepackage[normalem]{ulem}

\newcommand\XP[1]{\textcolor{black}{#1}}

\iccvfinalcopy % *** Uncomment this line for the final submission

\def\iccvPaperID{} % *** Enter the ICCV Paper ID here

\ificcvfinal\fi

\begin{document}

%%%%%%%%% TITLE
\title{Remind of the Past: Incremental Learning with Analogical Prompts}

\author{Zhiheng Ma$^1$, Xiaopeng Hong$^{2,3}$\footnotemark[1], Beinan Liu$^{4,1}$, Yabin Wang$^4$, Pinyue Guo$^5$, Huiyun Li$^1$\\
Advanced Technology, Chinese Academy of Science$^1$ Harbin Institute of Technology$^2$ \\
Peng Cheng Laboratory$^3$  Xi’an Jiaotong University$^4$  Jinan University$^5$\\
{\tt\small \{zh.ma,hy.li\}@siat.ac.cn hongxiaopeng@ieee.org} \\
{\tt\small \{pinna526,iamwangyabin\}@stu.xjtu.edu.cn guopinyue@stu2020.jnu.edu.cn}}


% \and
% Xiaopeng Hong\\

\maketitle
\ificcvfinal\thispagestyle{empty}\fi

\renewcommand{\thefootnote}{\fnsymbol{footnote}}
\footnotetext[1]{Corresponding author.}

\begin{abstract}
% Abstract goes here.

Although data-free incremental learning methods are memory-friendly, accurately estimating and counteracting representation shifts is challenging in the absence of historical data. This paper addresses this thorny problem by proposing a novel incremental learning method inspired by human analogy capabilities. Specifically, we design an analogy-making mechanism to remap the new data into the old class by prompt tuning. It mimics the feature distribution of the target old class on the old model using only samples of new classes. The learnt prompts are further used to estimate and counteract the representation shift caused by fine-tuning for the historical prototypes. 
The proposed method sets up new state-of-the-art performance on four incremental learning benchmarks under both the class and domain incremental learning settings. It consistently outperforms data-replay methods by only saving feature prototypes for each class. It has almost hit the empirical upper bound by joint training on the Core50 benchmark. The code will be released at \url{https://github.com/ZhihengCV/A-Prompts}.

\end{abstract}
\section{Introduction}
\label{sec:intro}

\begin{figure}[t] 
\centering
\includegraphics[width=0.95\linewidth]{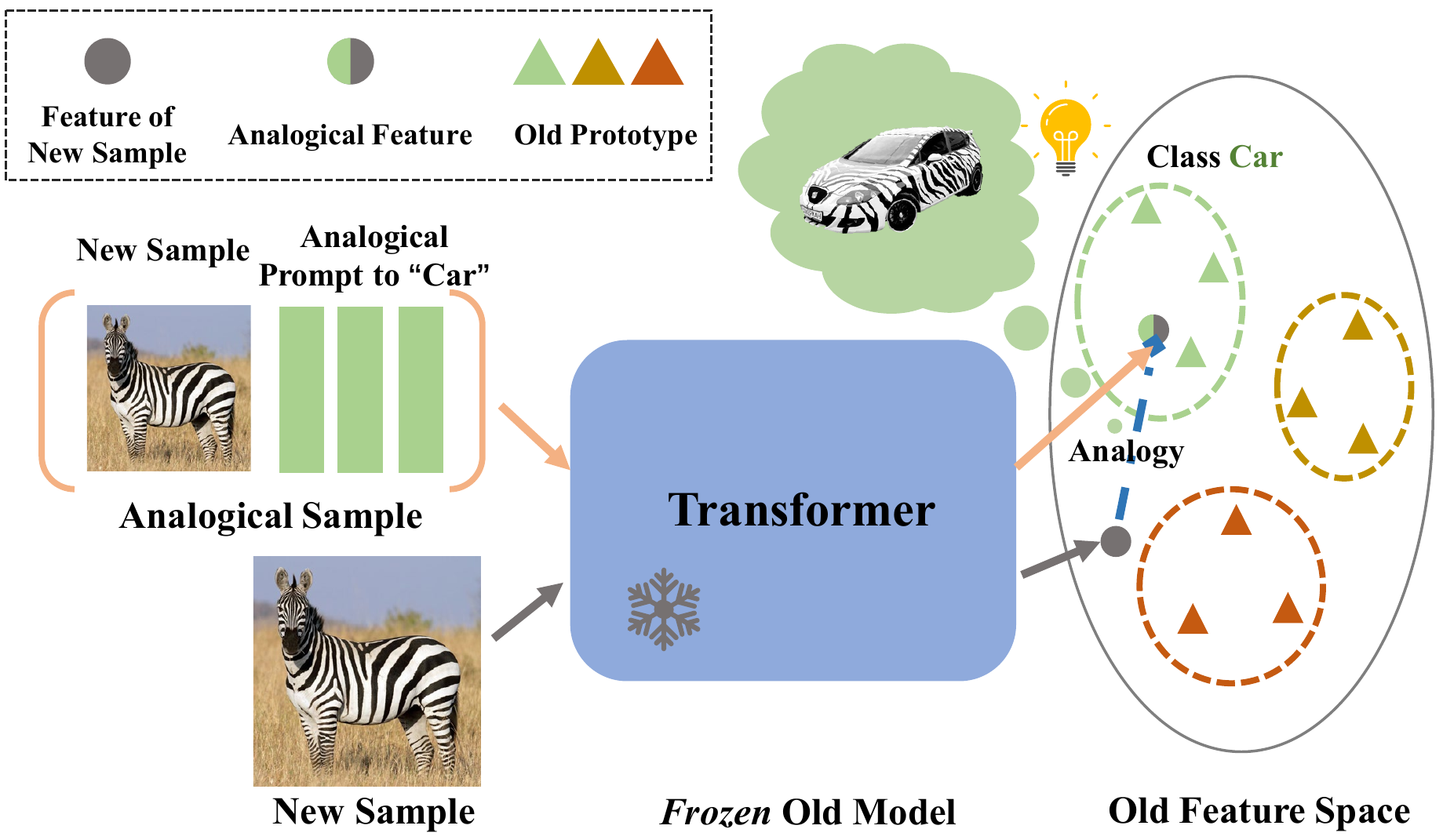} 
\caption{Illustration of the basic idea. Given a sample of the new class \emph{zebra}, the old model produces a feature vector far away from any previous class. We use the carefully designed prompts to guide the feature vector towards the old class \emph{car}, so that the frozen old model finally considers it as a \emph{car}. With such an analogy mechanism, for each new sample, we can find its counterpart feature vector to any category seen before (such as the one corresponding to the imaged \emph{car with stripes}), which is called the \emph{analogical feature}. The pairs of features of the \emph{analogical sample} by the \textbf{old} and the \textbf{updated new} models are further used to estimate \XP{and counteract} the representation shift caused by finetuning.
%for shift counteraction.
% the features by the \textbf{updated} model of the new samples and the corresponding analogical features by the \textbf{old} model are used to estimate the representation shift caused by the model update.
%With the help of carefully designed prompts, the feature vector is moving towards the old class \emph{car}, which the frozen old model finally considers as a \emph{car}.
} 
%, inevitably leading to more forgetting or interference
\label{fig:basicidea} 
% \vspace{-0.5cm}
\end{figure}

Incremental learning (IL), in which a single model is used to learn a non-stationary data sequence, is essential to achieve human-like AI. Many methods have been proposed to prevent \emph{catastrophic forgetting}~\cite{mccloskey1989catastrophic,french1999catastrophic} (the significant performance degradation of previously seen data) during incremental learning. A number of methods have been proposed, including  regularization-based~\cite{li2017LWF,kirkpatrick2017EWC,aljundi2018MAS}, architecture-based~\cite{yoon2017DEN,buzzega2020DER,mallya2018PackNet}, and replay-based~\cite{2017icarl,shin2017DGR} methods. 
Data-replay methods~\cite{lopez2017,chaudhryefficient, 2017icarl,rolnick2019ER,Hou_2019lucir,belouadah2019il2m} have superior performance in challenging scenarios, such as class-incremental learning (CIL). These methods use an external buffer to save representative data of previous tasks and replay them during the new task learning. Nonetheless, concerns have recently been raised about these data-replay methods~\cite{mai2022onlineSurvey,masana2020class}, as they are susceptible to memory constraints and cannot be used in scenarios with strict privacy protection.

Data-free-replay methods have been proposed to avoid explicitly saving data. Several methods~\cite{ostapenko2019learning,shin2017DGR,he2018exemplar,hu2019overcoming,xiang2019incremental,chenshen2018MemoryReplayGANs,kamra2017DGDM,NEURIPS2020GANMemory} adopt a separate generative model to generate pseudo data of previous tasks. However, maintaining and updating an additional generative model is computationally and memory intensive and still with privacy concerns~\cite{na2018theore}. Data-free knowledge distillation~\cite{yin2020dreaming} has been introduced into incremental learning~\cite{JamesSmith2021ABD,liu2022ERDR} to get rid of the additional generative model, which adopts the original classification model to synthesis the old images from scratch.  their performance is still lower than that of data-replay methods~\cite{JamesSmith2021ABD}. Another type of Data-free-replay methods~\cite{liu2020generative,zhu2022self,toldo2022bring,hayes2020remind,wang2021acae,zhu2021prototype,Petit_2023_WACV} stores the prototypes of each class in the feature space rather than the original images (feature-replay methods). These methods suffer from the representation shift, \emph{i.e.}, the features of the same input data change continuously as the model parameters are updated, making previously saved prototypes no longer applicable to the new feature space. Therefore, estimating the shift of the historical prototypes becomes critical. Previous methods~\cite{LuYu2020SDC,iscen2020memory} estimate the shift of the \emph{historical prototypes} by the feature shift of the \emph{current task's data}, which is biased and unreliable when the semantic gap between different tasks is significant (details elaborated in Sec.~\ref{sec:ptrain}). \textit{Without the old data, how can we fill the semantic gap and make an unbiased estimation of the representation shift?}

In this paper, we seek a new answer to this question from the inspiration of human intelligence. As a core mechanism of human intelligence, analogy is key for humans to learn and transfer knowledge efficiently to new domains. This ability enables us to connect different individual concepts, understand an infinite number of entities through associations and comparisons, and to continuously learn from experience~\cite{hofstadter2013Analogy}. People can easily find out the partial correspondence between different objects such as an \emph{armchair} is a chair with \emph{arms}.

% \MZ{ People can easily transform one entity to another with a simple \emph{prompt}. For example, a motorcycle is a \XP{bicycle powered by a motor (prompt); and an airport is a port (prompt)} for airplanes. (NOT Exactly)}

Inspired by this, we propose a novel data-free incremental learning approach to counteract the representation shift problem by explicitly building correspondence between the new and the old data. The basic idea is to imagine by analogy how a sample of a new class looks like in the old classes, which allows reminiscing about the old tasks while learning the new task. Once the corresponding feature vectors of the old class are generated using only the new samples, the semantic gap is resolved and we can calculate and then counteract the representation shift of the updated model in an unbiased manner. 

Specifically, our approach consists of three components. Firstly and most importantly, we propose a prompt-based analogy-making mechanism on top of the pre-trained vision transformer (ViT)~\cite{AlexeyDosovitskiy2020ViT}, as shown in Fig.~\ref{fig:basicidea}. We design a learning method with three carefully designed loss functions to obtain the \emph{analogical prompts} (A-prompts) for the classes seen before. On this basis, we can remap the samples of the new classes to the features of any previously-seen classes from the perspective of the old model, conditioned with the learnt A-prompts. For clarity and simplicity, the new samples conditioned with the learnt A-prompts and the corresponding feature vectors are named by the \emph{analogical samples} (A-samples) and the \emph{analogical features} (A-features), respectively. This mechanism is efficient as only the lightweight prompts need to be learnt, with the base model frozen.

Next, for an \emph{A-sample}, its corresponding \emph{A-feature} generated by the old model as well as the one generated by the updated model forms a pair. They are used to compute the instance-wise feature shift, and further accumulated to estimate the shift of the old class prototypes. On this basis, we counteract the shift of the old class prototypes at the new task and provide unbiased old prototypes for classification. Finally, we propose a tailored new task learning scheme to finetune the backbone model. Along with the classification loss, we propose a shift consistency loss to encourage the close points in the old feature space to have consistent shifts after finetuning, making the feature shift more predictable.

Our method is fundamentally different from previous prompt-based incremental learning methods, such as L2P~\cite{wang2022L2P}, DualP~\cite{wang2022DualP}, and S-Prompts~\cite{wang2022SPrompts}. These methods train task-specific prompts to reduce inter-task interference. In the test, they \XP{have to} perform a prompt-free inference to obtain the task ID \XP{first}, and then perform the task-prompt-conditioned inference to obtain the final prediction. In contrast \XP{to this two-pass inference, we only need a single pass of prompt-free inference, which significantly reduces, more specifically halves the inference cost of previous prompt-based methods. The main reason is that} our A-prompts are only used to update historical prototypes \XP{during training}, which are discarded upon completion. Moreover, our method eliminates the errors that arise from task ID prediction and further improves performance.

%Our method is fundamentally different from previous prompt-based incremental learning methods, such as L2P~\cite{wang2022L2P}, DualP~\cite{wang2022DualP}, and S-Prompts~\cite{wang2022SPrompts}. These methods train task-specific prompts to reduce inter-task interference. In the test, they first perform a prompt-free inference to obtain the task ID, and then perform the task-prompt-conditioned inference to obtain the final prediction. In contrast, our A-prompts are only used to update historical prototypes, which are discarded upon completion. In the test, we only need a single pass of prompt-free inference to obtain the result.  This reduces the inference cost to less than half of previous prompt-based methods and avoids the errors introduced by task ID prediction.

We evaluate the proposed method on four challenging  benchmarks, under both class incremental learning and domain incremental learning settings. Our method achieves state-of-the-art performance and even outperforms data-replay methods by only saving prototypes in feature space for each class. Moreover, our method \textit{has nearly touched the empirical upper bound} by joint training on Core50.
    
In summary, we propose a radically new incremental learning approach, which is inspired by the human analogy ability. The contributions of this paper are manifold:
\begin{itemize}

    \item We design a prompt-based analogy-making mechanism and the shift consistency loss for representation shift estimation and counteraction, which provides unbiased prototypes for classification.

    \item We propose an  analogical prompt learning method to produce the class-specific A-prompts, which plays a central role in linking the new samples to the old classes for analogy making.
    
    \item Our method reduces inference costs by half compared to previous prompt-based methods, and achieves state-of-the-art performance on four benchmarks.
\end{itemize}

\section{Related Work}
\label{sec:related}
% Overcoming catastrophic forgetting~\cite{mccloskey1989catastrophic,french1999catastrophic} has been a long-standing hotspot in incremental learning research. Existing methods can be roughly divided as follows.
%: 1) regularization-base, 2) memory-based, and 3) architecture-based methods. In addition, we present separately the methods that are most relevant to us, i.e., 4) data-free and 5) prompt-based methods.

\noindent\textbf{Continual Learning}. We only discuss the most related continual learning methods in this section, please refer to the latest surveys~\cite{masana2020class,wang2023comprehensive,de2021continual} for comprehensive introduction. \textit{Regularization-based methods} impose restrictions on the model's outputs or parameters. Output regularization methods~\cite{li2017LWF,2017icarl,Hou_2019lucir,wu2019bic,rannen2017encoder,castro2018EEIL,PrithvirajDhar2018LWM, ahn2021ss, kang2022class,douillard2020podnet,zhao2020maintaining} use knowledge distillation (KD)~\cite{hinton2015KD} to alleviate forgetting with the old model as the teacher.
Parameter regularization methods~\cite{kirkpatrick2017EWC,schwarz2018progress,aljundi2018MAS,ritter2018online,zenke2017SI,ahn2019uncertainty,chaudhry2018riemannian,liu2018rotate,lee2020continual} selectively constrain parameters according to their ``importance" on previous tasks. The regularization-based methods can further combine with other methods to further improve performance. \textit{Architecture-based methods}~\cite{pham2021dualnet,mallya2018PackNet,mallya2018Piggyback,liu2021adaptive,serra2018overcoming,shi2021continual} construct task-specific parameters to reduce interference between tasks, which includes neuron expansion~\cite{yoon2017DEN,xu2018reinforced,li2019learn}, network expansion~\cite{yan2021dynamically, wang2022foster, aljundi2017expert,schwarz2018progress,hung2019compacting}, and prompt expansion methods~\cite{douillard2022dytox,wang2022L2P,wang2022DualP,wang2022SPrompts}. Unlike prompt expansion methods, which use prompts to isolate different tasks, we use A-prompts to establish correspondence between different tasks. These prompts can be discarded after model updating, allowing our method to reason from original images without retrieving prompts from an expanding prompt pool. \textit{Data-replay methods}~\cite{2017icarl,rolnick2019ER,Hou_2019lucir,tao2020bocl,SonglinDong2021RKD,aljundi2019gradient,liu2020mnemonics} mitigate forgetting by keeping a buffer for the old data, which introduce the negative side effect of data imbalance~\cite{castro2018EEIL,wu2019bic,Hou_2019lucir,kim2020imbalanced,bang2021rainbow}. 

Our method can be categorized into the \textit{Data-free-replay methods}. Instead of generating pseudo samples of the previous task from scratch~\cite{shin2017DGR,chenshen2018MemoryReplayGANs,kamra2017DGDM,JamesSmith2021ABD,liu2022ERDR}, our method reprograms samples of the current task into the previous task, conditioned on the learnable prompts. Compared to previous feature-replay methods, our method has better plasticity because our feature representation is tunable rather than fixed~\cite{hayes2020remind,wang2021acae,Petit_2023_WACV}, and it better handles the representation shift problem~\cite{LuYu2020SDC,iscen2020memory} by addressing semantic gaps between different tasks. The continual learning ability of pretrained foundation models is also studied by~\cite{mehta2021empirical,wu2022class,ramasesh2022effect} with different setup.

\noindent\textbf{Prompt tuning}. Prompt tuning is first applied to natural language processing for efficient knowledge transfer of large foundation models~\cite{BrianLester2021PEPT,XiangLisaLi2021PrefixTuningOC,ZexuanZhong2021FactualPI,liu2023pre,houlsby2019parameter,raffel2020exploring,brown2020language,chenadaptformer}, which reprograms the input rather than finetunes the network. It is further applied to computer vision tasks, such as transfer learning~\cite{jia2022visual,bahng2022exploring,tsai2020transfer,zhou2022cocoop,zhou2022coop}, domain generalization~\cite{zheng2022prompt,gao2022visual}, and incremental learning~\cite{wang2022L2P,wang2022DualP,wang2022SPrompts}.
\section{Method}
\label{sec:method}
\newcommand{\argmax}{\mathop{\mathrm{argmax}}}
\newcommand{\x}{\boldsymbol{x}}
\newcommand{\f}{\boldsymbol{f}}
\newcommand{\p}{\boldsymbol{p}}
\newcommand{\A}{\boldsymbol{a}}
\newcommand{\B}{\boldsymbol{\varphi}}
\newcommand{\D}{\boldsymbol{\delta}}
\newcommand{\DD}{\boldsymbol{\Delta}}

\newcommand{\J}{\left[j\right]}
\newcommand{\lgt}{\boldsymbol{l}}
\newcommand{\old}{\boldsymbol{o}}
\newcommand{\new}{\boldsymbol{n}}
\newcommand{\para}{\boldsymbol{\theta}}
\newcommand{\pt}{t\!-\!1}
\newcommand{\pc}{\p_{y}}
\newcommand{\xc}{\tilde{\mathcal{X}}^{t}_{y}}

\subsection{Preliminaries and Notions}

\begin{figure}[!t] 
\centering
\includegraphics[width=0.95\linewidth]{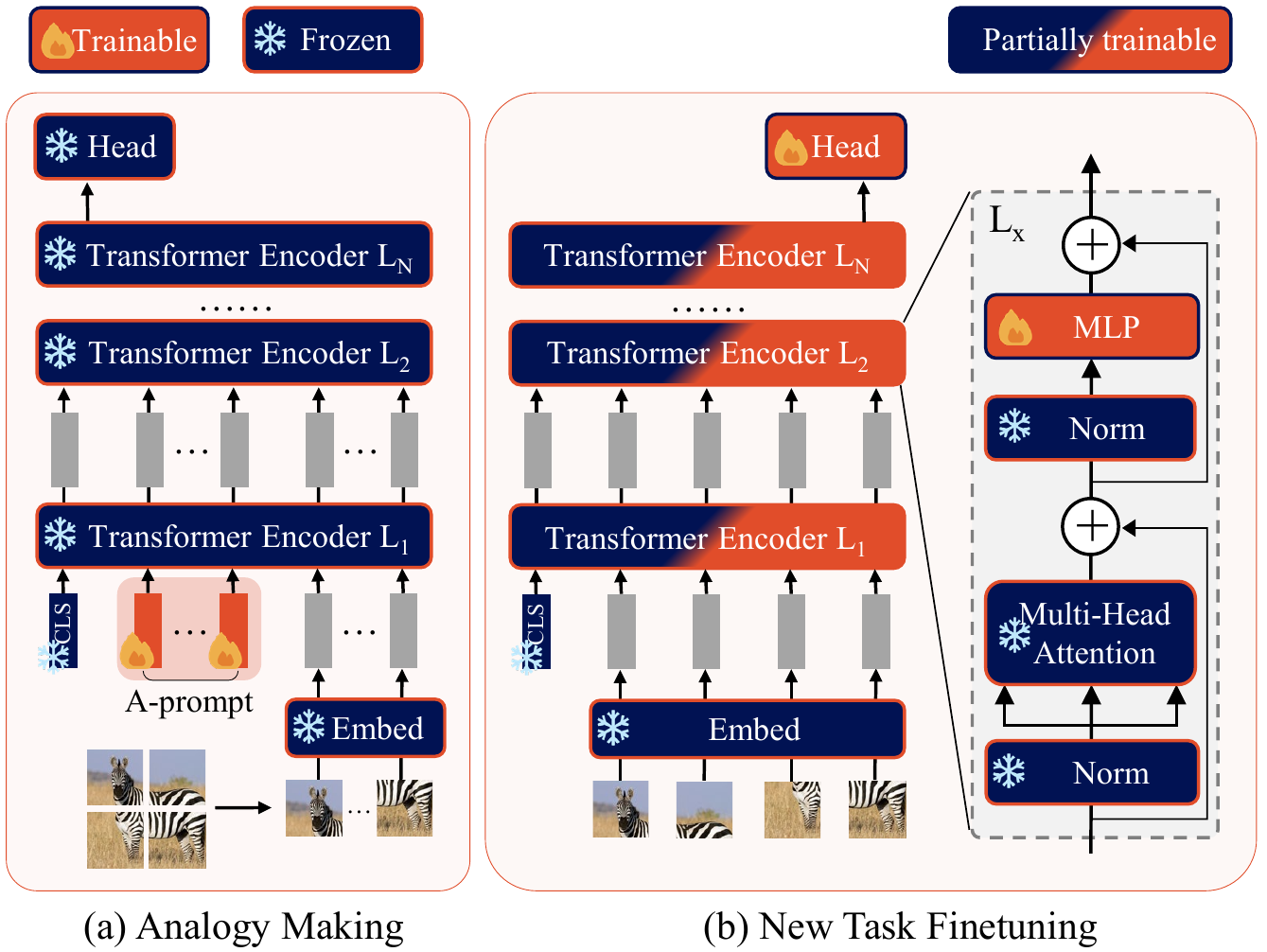} 
\caption{The network structure and the trainable parameters. We use the standard ViT as the backbone, and the training process includes two different stages: (a) In the analogy-making stage, we adopt the prompt-conditioned inference, and the only trainable parameter is the A-prompts; (b) In the new task finetuning stage, we use a workflow without prompt, where the trainable parameters are the MLP in each transformer encoder and the classification header.}
\label{fig:figure2}
\end{figure}

\noindent \textbf{Incremental Learning}. 
Incremental learning (IL) is to train a single model on a non-stationary data stream without catastrophic forgetting. In this paper, we focus on the scenario where a sequence of tasks is well defined, \emph{i.e.}, $t \in  \mathcal{T}=\left\{1,2,\cdots, T\right\}$, $T$ is the total task number, and we can only access the current task data. Let $\mathcal{Y}^{t}$ denote the label set of the task-$t$, and $\mathcal{S}^{t}$ denote all seen labels until the task-$t$,\emph{i.e.}, $\mathcal{S}^{t}$ = $\cup \left\{\mathcal{Y}^{1}, \mathcal{Y}^{2},...,\mathcal{Y}^{t} \right\}$. Let $\mathcal{X}^{t}$ represent the accessible samples of the task-$t$, and $(\x, y) \in \mathcal{Z}^{t}$ represent the sample-label pairs of the task-$t$. $\mathcal{X}_y$ represents samples of the class-$y$. 

\noindent \textbf{Vision Transformer and Prompt Tuning}.
The vision transformer (ViT) backbone consists of two components, \emph{i.e.}, $f = f_a \circ f_e$, where $f_e$ is the embedding layer which transforms the input image $\x \in \mathbb{R}^{H \times W \times C}$ into patch tokens $f_e(\x) \in \mathbb{R}^{(L+1) \times D}$, whose first token is the $[class]$ token inherited from the pre-trained model; $f_a$ consists of a sequence of transformer encoders, each of which includes a multi-head attention (MHA) and a multi-layer perception (MLP). We take the first token $\f \in \mathbb{R}^{D}$ of the last layer, the corresponding output of the $[class]$ token, as the feature vector for classification tasks, \emph{i.e.}, $\f = f_a(f_e(\x))\left[0\right]$.

As $f_a$ allows the arbitrary length of tokens as input, we can attach additional learnable tokens to the original image tokens, \emph{i.e.}, $f_e(\x) \oplus \p$. $\oplus$ denotes the concatenate operator along the token sequence; $\p \in \mathbb{R}^{J \times D}$ is the learnable prompt token, which has the same feature dimension as the image tokens. Finally, the prompt-conditioned feature vector can be calculated as $\tilde{\f}= f_a(f_e(\x) \oplus \p)\left[0\right]$, $\tilde{\f} \in \mathbb{R}^{D}$. In the following section, we abbreviate the above two equations:
\begin{equation}
  \f = f_{\para}(\x), \tilde{\f} = f_{\para}(\x|\p),  
\end{equation}
where $\f$ and $\tilde{\f}$ are the original and the prompt-conditioned feature vector, respectively. $\para$ is the model's parameters. Furthermore, $f^{\pt}(\cdot)$ is the abbreviation for $f_{\para^{\pt}}(\cdot)$, which represents the model before the task-$t$ finetuning; $f^{t}(\cdot)$ is the abbreviation for $f_{\para^{t}}(\cdot)$, which represents the model after the task-$t$ finetuning. More details about ViT and prompt tuning can be found in~\cite{AlexeyDosovitskiy2020ViT,liu2023pre}.

\subsection{Overall Framework}

We propose a Soft-Nearest-Multi-Prototype (SNMP) classifier in our method, which is an extension of  the Nearest-Mean-of-Exemplars (NME) Classifier~\cite{2013distance}. NME is widely used in incremental learning~\cite{2013distance,2017icarl,LuYu2020SDC} since it can easily add novel classes by calculating their mean centers in feature space, and less bias to the latest task. Our SNMP maintains multiple prototypes for each class instead of a single mean center, better characterizing the feature distribution and the complex boundaries between different classes, which is formulated as follows: 
\begin{equation}\label{eq:class}
    y^* = \argmax_{y \in \mathcal{S}^{t}} \frac{\sum_{m=1}^{M} e^{-d(f_{\para}(\x), \B_{ym})}}{\sum_{y \in \mathcal{S}^{t}}\sum_{m=1}^{M} e^{-d(f_{\para}(\x), \B_{ym})}},
%y^* = \argmax_{y \in \mathcal{S}^{t}} \frac{\sum_{m=1}^{M} \exp(\beta \cos(f_{\para}(\x), \B_{ym}))}{\sum_{y \in \mathcal{S}^{t}}\sum_{m=1}^{M} \exp(\beta \cos(f_{\para}(\x), \B_{ym}))}   
\end{equation}
where $\B_{ym}$ represents the $m$-th prototype of the class $y$, and $M$ is the total prototype number per class. $d(\cdot,\cdot)$ can be any distance function between two vectors. 
% $\beta$ is the scaling factor that controls the smoothness of classification boundaries. 
We abbreviate $\B_{ym}$ in the following as $\B_{y}$ without ambiguity because constructing and maintaining multiple prototypes is just a simple repetition of one prototype.

We can easily construct prototypes of the current task by taking the K-Means clustering centers for each new class. The main difficulty is how to keep track of historical prototypes affected by the representation shift. data-replay methods~\cite{2013distance,2017icarl} can simply re-inference the saved old data on the new model to obtain unbiased prototypes, which is not possible in the data-free setting. Therefore, we propose the analogical prompts (A-prompts) which ``convert" the new data into old classes and use the analogical features (A-features) to obtain the shift-counteracted prototypes. We name the new sample conditioned with the A-prompts as analogical samples (A-samples).

In the following subsections, we will elaborate on how to obtain the shift-counteracted prototype $\B_{y}$ (Sec.~\ref{sec:ptrain}) and efficiently learn the ViT backbone $f_{\para}(\x)$ (Sec.~\ref{sec:finetune}).

\subsection{Analogical Shift Counteraction}\label{sec:ptrain}

\begin{figure}[!t] 
\centering
\includegraphics[width=0.95\linewidth]{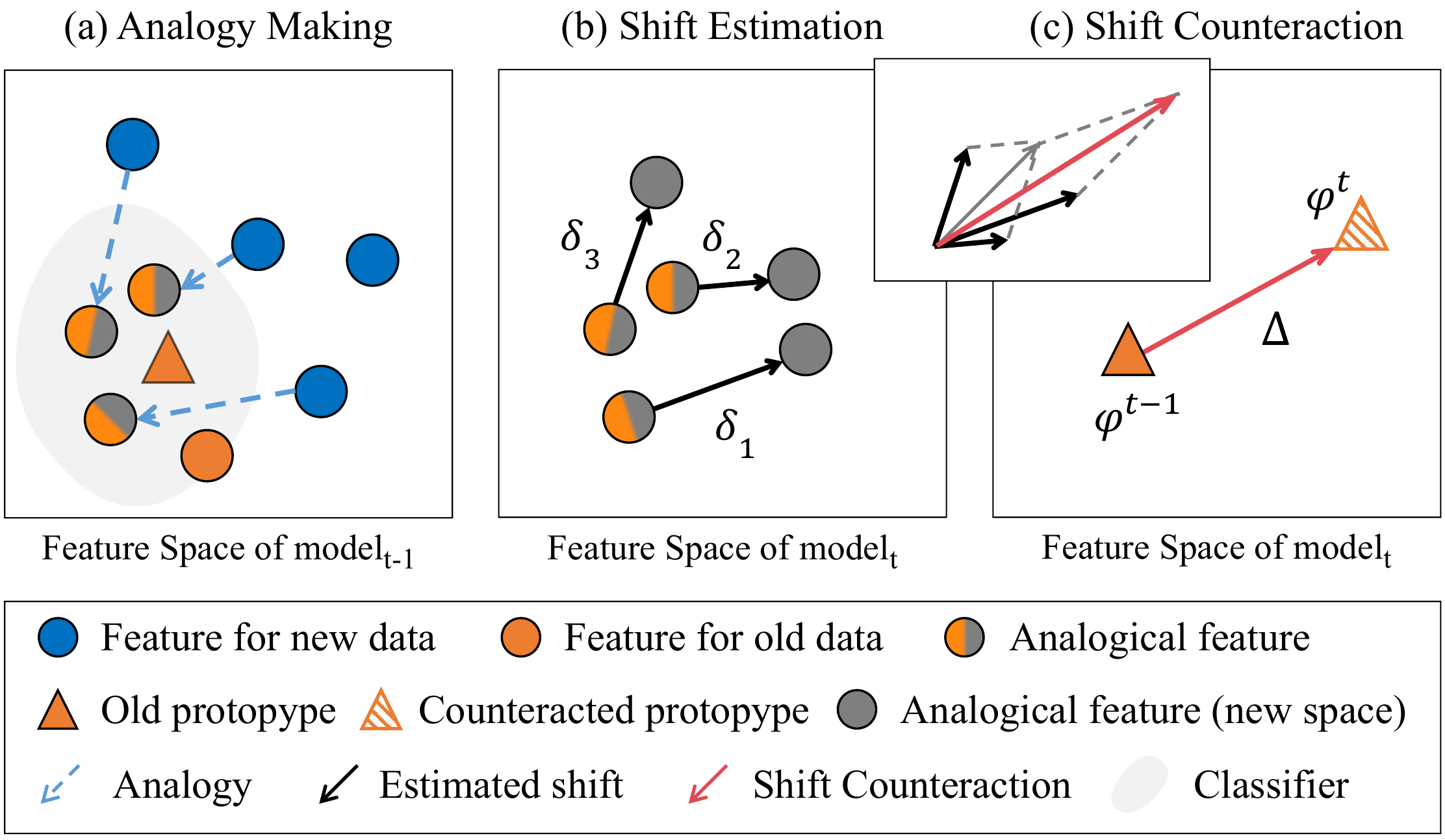} 
\caption{Analogical shift counteraction. (a) We push the K-Nearest new samples towards the old prototype and make it be classified into the old class from the perspective of the old model with the learned class-specific A-prompts; (b) We obtain the instance-wise shift between the pair of analogical features on the new and the old model; (c) We counteract the old prototype's shift in the updated feature space by adding the weighted average shift of its reference analogical features.
}
\label{fig:figure3}
\end{figure}

\noindent \textbf{Analogy Making}. In order to retain old knowledge and resist forgetting, the prompt-based analogy mechanism makes the A-samples behave like the old samples from the perspective of the old model. We can consider it from following two aspects: first, the A-samples will be classified into the target old class on the old model; second, A-features should follow the corresponding distribution in the old feature space. By meeting the above two criteria, the A-samples can substitute the old samples to some extent, which is further used to counteract the prototype shift.

We describe the training process of our A-prompts in the following. To accelerate the training process, we first select the most similar subset from the current task data $\mathcal{X}^{t}$ according to their distance to the target prototype $\B^{\pt}_y, y \in \mathcal{S}^{\pt}$:
\begin{equation}\label{eq:sample}
    \xc = \left\{ \x | \x \in \mathcal{X}^{t}, f^{\pt} (\x) \in \text{K-NN}(\B^{\pt}_y) \right\}, 
\end{equation}
where $K$ is the total sample number for each historical prototype, and $\text{K-NN}(\cdot)$ is the K-Nearest-Neighbor of a given feature vector. We use this subset to train the class-specific prompts $\left\{\pc | y \in \mathcal{S}^{\pt}\right\}$. In the prompt training stage, the only trainable parameters are A-prompts while the old model is frozen (visualized in Fig.~\ref{fig:figure2} (a)). We propose an objective function that includes three components:
\begin{equation}\label{eq:pt}
 \mathcal{L}_{PT} = \mathcal{L}_{CC} + \mathcal{L}_{PP} + \mathcal{L}_{DE}, \x \in \xc. 
\end{equation}

\noindent The first component is \XP{the} class-convert (CC) loss to remap the A-samples into the target old class $y \in \mathcal{S}^{\pt}$: 
\begin{equation}\label{eq:adv}
        \mathcal{L}_{CC} = -\frac{1}{N}\sum_{i=1}^{N} \log(h^{\pt}(\tilde{\f}^{\pt}_{i|y})\left[ y \right]),
\end{equation}
where $N$ is the training batch size; $\tilde{\f}_{i|y}^{\pt} = f^{\pt}(\x_i| \pc)$ is the $i$-th A-feature of the class-$y$ from the perspective of the old model; $h(\cdot)$ is the fully-connected classification layer with the softmax normalize function.

The second component is named prototype-push (PP) loss, which is designed to push the A-features towards to the target prototype: 
\begin{equation}\label{eq:pp}
    \mathcal{L}_{PP} = \frac{1}{N}\sum_{i=1}^{N}d(\B_{y}^{\pt}, \tilde{\f}_{i|y}^{\pt}). 
\end{equation}

The third component is named diversity-encourage (DE) loss, which prevents A-features from collapsing into a single point:
\begin{equation}\label{eq:de}
    \mathcal{L}_{DE} = \frac{1}{N(N-1)}\sum_{i=1}^{N}\sum_{j=i+1}^{N}\lfloor \Omega - d(\tilde{\f}_{i|y}^{\pt}, \tilde{\f}_{j|y}^{\pt}) \rfloor_{+},
\end{equation}
where $\lfloor \cdot \rfloor_{+}$ truncates negative value to zero; $\Omega$ is a constant margin, which is set to 1 in our experiments. 

Once the A-prompts are obtained, we can counteract the shift of the historical prototypes, which is described below.

\begin{figure}[!t] 
\centering
\includegraphics[width=0.7\linewidth]{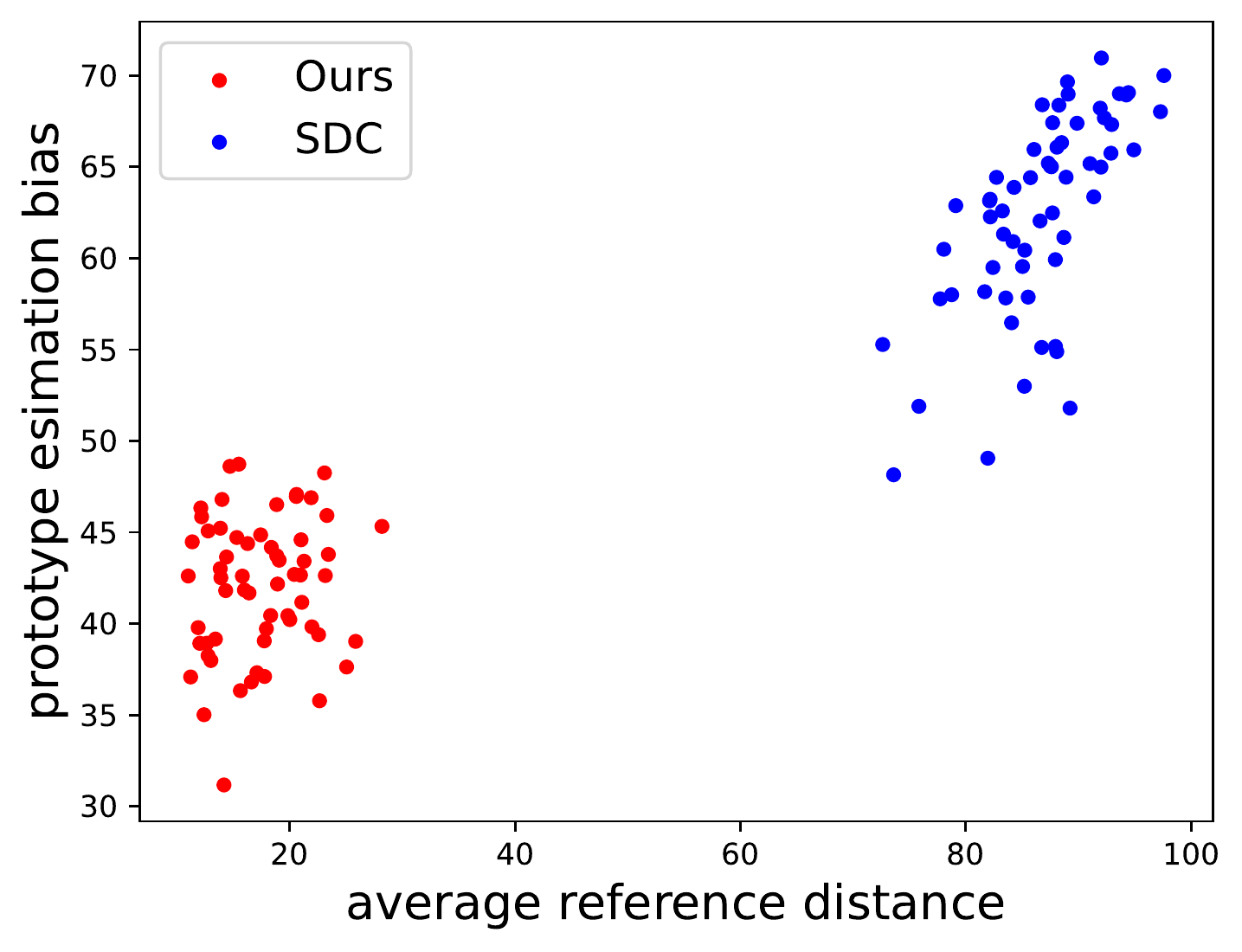}
\caption{Comparison of prototype estimation bias. The prototype estimation bias is the distance between the shift-counteracted prototypes and the ground-truth prototypes after the new task finetuning; the average reference distance is the average distance between the reference features and the prototypes before finetuning, \emph{i.e.}, $d(\tilde{\f}^{\pt}_{i|y},\B_{y}^{\pt})$ for ours and $d(\f^{\pt}_i,\B_{y}^{\pt})$ for SDC, respectively.}\label{fig:bias}
\end{figure}

\noindent \textbf{Shift Estimation and Counteraction}.
The pairs of features of the A-samples by the old and the updated new models are used to estimate the representation shift caused by finetuning:
\begin{equation}\label{eq:shift}
    \D_{i|y} = \tilde{\f}^{t}_{i|y} - \tilde{\f}^{\pt}_{i|y},
\end{equation}
where $\tilde{\f}_{i|y}^{t} = f^{t}(\x_i| \pc)$ is the $i$-th A-feature of the class-$y$ in the new feature space, $\D_{i|y}$ is the instance-wise shift between the new and the old A-feature. Since features close together in the old feature space should have similar shifts after finetuning, we can estimate the prototype shift as follows:
\begin{equation}\label{eq:cps_p}
    \DD_y = \frac{\sum_{i} \exp( -d(\tilde{\f}^{\pt}_{i|y}, \B^{\pt}_{y})) \D_{i|y}}{\sum_{i} \exp( -d(\tilde{\f}^{\pt}_{i|y}, \B^{\pt}_{y}))},
\end{equation}
where $\DD_y$ represents the estimated shift of prototype $\B^{\pt}_{y}$. Finally, we can counteract the shift of old prototypes as follows:
\begin{equation}\label{eq:counteract}
    \B^{t}_{y} = \B^{\pt}_{y} + \DD_{y}.
\end{equation}

The diagram of the analogical shift counteraction process is shown in Fig.~\ref{fig:figure3}.

\noindent \textbf{The rationality of using A-features:} SDC~\cite{LuYu2020SDC} also uses the shift counteraction to update historical prototypes. Nevertheless, it directly takes the original features of new data as the reference rather than the A-features. That is, in Eq.~\eqref{eq:shift} and Eq.~\eqref{eq:cps_p}, SDC replaces $\tilde{\f}^{t}_{i|y}$ and $\tilde{\f}^{\pt}_{i|y}$ with $\f^{t}_{i}$ and $\f^{\pt}_{i}$, respectively, where $\f_{i}^{t} = f^{t}(\x_i), \x_i \in \mathcal{X}^{t}$. The intuitive assumption behind Eq.~\eqref{eq:cps_p} is that the closer the points are in the feature space, the more consistent the shifts will be after the finetuning.  Therefore, the reliability and accuracy of the prototype shift counteraction is highly dependent on how close the reference features are to the prototype. We verify this assumption with analytical experiments. Fig.~\ref{fig:bias} visualizes relationship between the average reference distance and shift estimation bias. As can be seen, larger average reference distances result in more significant prototype estimation bias for SDC. In contrast, Our method has less prototype estimation bias than SDC, since our A-features have already been pushed towards to their target prototypes by Eq.~\eqref{eq:pp}. The TSNE~\cite{van2008visualizing} visualization of the feature space can be found in the supplementary material. Finally, our method achieves consistent performance improvement over SDC (Sec.~\ref{sec:compare}).

\subsection{New Task Finetuning}\label{sec:finetune}

In this section, we describe how to finetune the old model $f^{\pt}$ on the task-$t$ to get the new model $f^{t}$. For training efficiency and alleviating catastrophic forgetting, we only update the transformer encoder's MLP layer and the classification header, as visualized in Fig.~\ref{fig:figure2} (b). The training objective of the task-$t$ is defined as follows:
\begin{equation}\label{eq:ft}
 \mathcal{L}_{FT} = \mathcal{L}_{C} +  \mathcal{L}_{SC}, \x \in \mathcal{X}^{t}, 
\end{equation}
where $\mathcal{L}_{C}$ is the cross-entropy loss on the current task~\cite{castro2018EEIL}; $\mathcal{L}_{SC}$ is the shift consistency loss, which encourages local feature shifts to become consistent:
\begin{equation}
    \begin{aligned}
            \mathcal{L}_{SC} &= \frac{1}{N}\sum_{i=1}^{N} d\left(\D_i, \DD_i\right), \D_{i} = \f_{i} - \f^{\pt}_{i},\\
            \DD_{i} &= \frac{\sum\limits_{j,j \neq i} \exp( -d(\f^{\pt}_{j}, \f^{\pt}_{i})) \D_{j}}{\sum\limits_{j,j \neq i} \exp( -d(\f^{\pt}_{j}, \f^{\pt}_{i}))},
    \end{aligned}    
\end{equation}
where $\D_{i}$ is the true feature shift of the $i$-th sample during finetuning, $\DD_{i}$ is the feature shift estimated by its nearby feature shifts.

It is worth noting that we only use the new samples $\mathcal{X}^{t}$ to finetune our model, without using A-prompts to ``analogize" samples of old classes, because we find that distinguishing A-samples from original samples is easy to over-fitting, and destroys the prompt-free test accuracy. After finishing the model finetuning on the new task, we can easily initialize new prototypes by taking the K-Means clustering centers for each new class and counteracting the shift of old prototypes by Eq.~\eqref{eq:counteract}. Finally, we can classify the test samples with Eq.~\eqref{eq:class}. More details about our training process can be found in Alg.~\ref{alg:algo_train}.

\begin{algorithm}
\caption{Model Training}\label{alg:algo_train}
\begin{footnotesize}
\KwIn{the total task number $T$; the pre-trained ViT model $f$; the training data sequence $\left\{\mathcal{Z}^{t}|t \in \mathcal{T}\right\}$}
\KwOut{$f^{T}$; $\left\{\B^{T}_{y}|y \in \mathcal{S}^{T}\right\}$}
Initialize $f^{0}=f$;\\
\tcp{Start from the pre-trained model}
\For{$t=1,\ldots,T$}{
\eIf{$t=1$}{ Train $f^{\pt}$ by $\mathcal{L}_{C}$ on $\mathcal{Z}^{t}$ to get $f^{t}$; \\
             \tcp{Finetune on the first task}
             Get $\left\{\B^{t}_{y}|y \in \mathcal{Y}^{t}\right\}$ by class-wise K-Means; \\
             \tcp{Initialize prototypes}
}{
    \For{$y=1,\ldots,|\mathcal{S}^{\pt}|$}{
    \tcp{Repeat on all old classes}
    Sample $\xc$ by Eq.~\eqref{eq:sample}; \\
    \tcp{Samples for training A-prompt}
    Train $\pc$ by Eq.~\eqref{eq:pt} on $\xc$; \\
    \tcp{Train class-specific A-prompt}
    }
    Training $f^{\pt}$ by Eq.~\eqref{eq:ft}  on $\mathcal{Z}^{t}$ to get $f^{t}$; \\
    \tcp{Finetune model on the new task}
    Get $\left\{\B^{t}_{y}|y \in \mathcal{Y}^{t}\right\}$ by class-wise K-Means; \\
    \tcp{Construct new prototypes}
    Update $\left\{\B^{\pt}_{y}|y \in \mathcal{S}^{\pt}\right\}$ to $\left\{\B^{t}_{y}|y \in \mathcal{S}^{\pt}\right\}$ in-place by Eq.~\eqref{eq:counteract} with $\left\{(\pc, \xc)|y \in \mathcal{S}^{\pt}\right\}$;
    \tcp{Update old prototypes}
}
abandon $f^{\pt}$, $\mathcal{Z}^{t}$, $\left\{(\pc, \xc)|y \in \mathcal{S}^{\pt}\right\}$;
}
\Return $f^{T}$, $\left\{\B^{T}_{y}|y \in \mathcal{S}^{T}\right\}$
\end{footnotesize}
\end{algorithm}

\section{Experiments}
\label{sec:experiments}
As our method is built upon the pre-trained ViT  model, we closely follow the setting proposed by previous state-of-the-art ViT-based methods such as L2P~\cite{wang2022L2P} and DualP~\cite{wang2022DualP}. We mainly focus on the challenging class incremental learning and domain incremental learning settings, where the task ID is not available during inference. 

\subsection{Benchmark and Implementation}
\noindent \textbf{Benchmark Datasets.} 
We conduct experiments on four challenging benchmark datasets: Split CIFAR-100~\cite{AlexKrizhevsky2009cifar}, Split ImageNet-R~\cite{wang2022DualP}, Sequential 5-Datasets~\cite{SaynaEbrahimi20205datasets}, and Core50~\cite{VincenzoLomonaco2017core50}. \textit{Split CIFAR-100} splits the original CIFAR-100~\cite{AlexKrizhevsky2009cifar} into 10 sessions, each containing 10 disjoint classes. \textit{Split ImageNet-R} is a challenging class incremental benchmark. It randomly divides the original ImageNet-R's~\cite{DanHendrycks2020imagenetr} 200 classes into 10 sessions, each with 20 classes. It is usually used for incremental learning with pre-trained models. \textit{Sequential 5-Datasets} incorporates five popular datasets, which are CIFAR-10, MNIST~\cite{lecun1998mnist}, Fashion-MNIST~\cite{xiao2017fashionmnist}, SVHN~\cite{netzer2011svhn}, and notMNIST~\cite{bulatov2011notmnist}. The combined datasets provide greater diversity for the evaluation of incremental learning approaches. \textit{Core50} is a trendy domain incremental dataset that contains 11 distinct domains with 50 classes each. Among them, 8 domains are used in the training phase, and 3 for the testing phase. Following the setup of L2P~\cite{wang2022L2P} and DualP~\cite{wang2022DualP}, we use Split CIFAR-100, Split ImageNet-R, and Sequential 5-Datasets for class incremental learning experiments, and use Core50 for domain incremental learning experiments.

\noindent \textbf{Implementation Details.} 
For a fair comparison, \textit{all methods} start with the same ImageNet pre-trained ViT-B/16~\cite{AlexeyDosovitskiy2020ViT}, with the hyper-parameters from the original setting. 

For the \emph{new task finetuning} stage, we use SGD optimizer with the learning rate = 0.001, the momentum = 0.9, and the batch size = 128. The training epoch is set to 5 for Split CIFAR-100 and Sequential 5-Datasets, 10 for Core50, and 50 for Split ImageNet-R, which strictly follows previous works~\cite{wang2022L2P,wang2022DualP} for a fair comparison. The \emph{trainable} parameters are the MLP in each transformer encoder and the classification header.

For the \emph{analogy making} stage, we use Adam optimizer with the learning rate = 0.001. The training epoch is set to 5 for Split CIFAR-100, ImageNet-R, and Core50, and 10 for Sequential 5-Datasets, according to the level of semantic gap between different tasks. 5-Datasets has the largest semantic gap between tasks among these four benchmarks. The \emph{trainable} parameters are A-prompts. 

For the \emph{hyper-parameters setting}, except for the hyper-parameters study experiments, we set the total prototypes per class $M$ to $6$, with comparable additional parameters to the DualP; We take scaling normalized euclidean distance as the distance function $d(\f_i,\f_j)=20 \cdot \ell_2\left(\frac{ \f_i}{\Vert\f_i\Vert_{2}}, \frac{ \f_j}{\Vert\f_j\Vert_{2}}\right)$, where the scaling factor is set to 20; we set $K$ to 50 in Eq.~\eqref{eq:sample}, and the token length $J$ of each prompt to 5.  

We evaluate our method under both the class incremental learning (CIL) and domain incremental learning (DIL) settings.
In DIL where different domains share the same categories, we directly use A-prompts to convert samples to the identity category in different domains, without the need to select samples by Eq.~\eqref{eq:sample}. The other processes are the same as in CIL. 

\noindent \textbf{Evaluation Metrics.} 
We use the Final Average Accuracy (FAA) and Final Forgetting (FF) as evaluation metrics~\cite{wang2022DualP,wang2022L2P,wang2022SPrompts}, both of which evaluate the model after the last task. FAA is more important than FF, since FAA is related to both the plasticity and stability of the model, while FF is only related to the stability of the model.

\subsection{Comparison and Analysis}\label{sec:compare}

We compare our method with eleven modern CIL and DIL methods, including: 
1) Prompt-based methods, which achieve the top accuracy in the ViT-based incremental learning, including L2P~\cite{wang2022L2P}, DualP~\cite{wang2022DualP}, and S-Prompts~\cite{wang2022SPrompts}.
2) Data-replay methods, which use a buffer to save representative historical data, including ER~\cite{rolnick2019ER}, BiC~\cite{wu2019bic}, GDumb~\cite{prabhu2020gdumb}, DER++~\cite{PietroBuzzega2020der+}, and Co$^2$L~\cite{HyuntakCha2021co2l}. Since they are sensitive to the buffer size, a medium and a large buffer size are reported as ~\cite{wang2022L2P,wang2022DualP}. 3) Weight-regularization methods including LwF~\cite{li2017LWF} and EWC~\cite{kirkpatrick2017EWC}, which are widely used baseline methods. All these methods are re-implement on ViT for a fair comparison and the results are taken from~\cite{wang2022DualP,wang2022SPrompts,wang2022L2P}
%These latest state-of-the-art methods are re-implement on ViT for fair comparison.

%All results are taken from~\cite{wang2022DualP,wang2022SPrompts} except for our method, our baseline methods, and the upper-bound result. 
We also compare with SDC-SNMP~\cite{LuYu2020SDC}  to illustrate the effectiveness of our main contribution, \emph{i.e.}, the analogical shift counteraction. It uses the same classifier (Eq.~\eqref{eq:class}) and the same finetuning objective function (Eq.~\eqref{eq:ft}) as ours. Moreover, we train a model on the entire training set in a batch manner with the same trainable parameters as ours, which has the highest performance on the same architecture. It is considered as the \emph{upper bound} in the experiment.

The experimental results are shown in Tabs.~\ref{table:cifar} to~\ref{table:core50}. 
%The highlights and the conclusions are listed as follows: 
%improves the best performance
1) Our method \XP{performs best} on Split CIFAR-100, Split Image-R, and Core50, and achieves a performance with a marginal difference to the best one on 5-Datasets. 2) For DIL on Core50, our method achieves a performance very close to the \emph{upper bound}. 3) Our method outperforms all the data-replay methods with much lower memory costs.

\noindent \textbf{Comparison with SDC-SNMP}: our method consistently outperforms SDC-SNMP on all four datasets. It strongly supports the effectiveness and necessity of analogical shift counteraction. It is worth noting that the performance margin between SDC and our method is significant when the semantic gap between different tasks is large and vice versa. For instance, on the most diverse benchmark Sequential 5-Datasets, our method outperforms SDC-SNMP by $7.12\%$.

% ################################################
% 【split cifar-100】
% ################################################
\begin{table}[] \small
\setlength{\tabcolsep}{5pt}
\begin{center}
\begin{tabular}{lccc}
\hline
Method & Buffer size & FAA ($\uparrow$) & FF ($\downarrow$) \\ 
\hline
ER~\cite{rolnick2019ER} & \multirow{6}{*}{10/class} &
    67.87\scriptsize{$\pm$0.57} & 
    33.33\scriptsize{$\pm$1.28}  
    \\
BiC~\cite{wu2019bic} && 
    66.11\scriptsize{$\pm$1.76} & 
    35.24\scriptsize{$\pm$1.64}  
    \\
GDumb~\cite{prabhu2020gdumb} && 
    67.14\scriptsize{$\pm$0.37} & 
    - \\
DER++~\cite{PietroBuzzega2020der+} && 
    61.06\scriptsize{$\pm$0.87} & 
    39.87\scriptsize{$\pm$0.99}  
    \\
Co$^2$L~\cite{HyuntakCha2021co2l} && 
    72.15\scriptsize{$\pm$1.32} & 
    28.55\scriptsize{$\pm$1.56}  
    \\
L2P~\cite{wang2022L2P} &&
    84.21\scriptsize{$\pm$0.53} &
    7.72\scriptsize{$\pm$0.77}
    \\
\hline
ER~\cite{rolnick2019ER} & \multirow{6}{*}{50/class} &
    82.53\scriptsize{$\pm$0.17} & 
    16.46\scriptsize{$\pm$0.25}  
    \\
BiC~\cite{wu2019bic} && 
    81.42\scriptsize{$\pm$0.85} & 
    17.31\scriptsize{$\pm$1.02}  
    \\
GDumb~\cite{prabhu2020gdumb} && 
    81.67\scriptsize{$\pm$0.02} & 
    - \\
DER++~\cite{PietroBuzzega2020der+} && 
    83.94\scriptsize{$\pm$0.34} & 
    14.55\scriptsize{$\pm$0.73}  
    \\
Co$^2$L~\cite{HyuntakCha2021co2l} && 
    82.49\scriptsize{$\pm$0.89} & 
    17.48\scriptsize{$\pm$1.80}  
    \\ 
L2P~\cite{wang2022L2P} &&
    86.31\scriptsize{$\pm$0.59} &
    5.83\scriptsize{$\pm$0.61}
    \\
\hline
FT-seq & \multirow{6}{*}{0} & 
    33.61\scriptsize{$\pm$0.85} & 
    86.87\scriptsize{$\pm$0.20}  
    \\
EWC~\cite{kirkpatrick2017EWC} && 
    47.01\scriptsize{$\pm$0.29} & 
    33.27\scriptsize{$\pm$1.17}  
    \\
LwF~\cite{li2017LWF} && 
    60.69\scriptsize{$\pm$0.63} &
    27.77\scriptsize{$\pm$2.17}
    \\
L2P~\cite{wang2022L2P} && 
    83.86\scriptsize{$\pm$0.28} & 
    7.35\scriptsize{$\pm$0.38}  
    \\
DualP~\cite{wang2022DualP} && 
    86.51\scriptsize{$\pm$0.33} & 
    5.16\scriptsize{$\pm$0.09} 
    \\ 
\cdashline{1-4}[0.8pt/2pt]
% PT-SNMP & \multirow{4}{*}{0} & 
%     75.37\scriptsize{$\pm$0.00} & 
%     7.43\scriptsize{$\pm$0.00} 
%     \\ 
% FTF-SNMP && 
%     84.78\scriptsize{$\pm$0.00} & 
%     4.64\scriptsize{$\pm$0.00} 
%     \\ 
SDC-SNMP &\multirow{2}{*}{0}& 
    82.31\scriptsize{$\pm$0.57} & 
    14.44\scriptsize{$\pm$0.68} 
    \\
Ours && 
    \bf{87.87\scriptsize{$\pm$0.24}} & 
    \bf{2.78\scriptsize{$\pm$0.07}}  
    \\
% Baseline4 && 
%     73.13\scriptsize{$\pm$0.00} & 
%     10.59\scriptsize{$\pm$0.00} 
%     \\ 
\hline
\hline
Upper bound & - & 
    91.98\scriptsize{$\pm$0.07} &
    - \\ 
    \hline
\end{tabular}
\end{center}
\caption{Result on Split CIFAR-100 for CIL. %Memory-based methods get results at buffer sizes of 50/class and 10/class. 
Buffer size 0 means no buffer required.
%\textbf{Bold}: the best data-free result.
}
\label{table:cifar}
\end{table}

% ################################################
% 【split ImageNet-R】
% ################################################

\begin{table}[] \small
\setlength{\tabcolsep}{5pt}
\begin{center}
\begin{tabular}{lccc}
\hline
Method & Buffer size & FAA ($\uparrow$) & FF ($\downarrow$) \\ 
\hline
ER~\cite{rolnick2019ER} & \multirow{6}{*}{10/class} &
    55.13\scriptsize{$\pm$1.29} & 
    35.38\scriptsize{$\pm$0.52}  
    \\
BiC~\cite{wu2019bic} && 
    52.14\scriptsize{$\pm$1.08} & 
    36.70\scriptsize{$\pm$1.05}  
    \\
GDumb~\cite{prabhu2020gdumb} && 
    38.32\scriptsize{$\pm$0.55} & 
    - \\
DER++~\cite{PietroBuzzega2020der+} && 
    55.47\scriptsize{$\pm$1.31} & 
    34.64\scriptsize{$\pm$1.50}  
    \\
Co$^2$L~\cite{HyuntakCha2021co2l} && 
    53.45\scriptsize{$\pm$1.55} & 
    37.30\scriptsize{$\pm$1.81}  
    \\ 
\hline
ER~\cite{rolnick2019ER} & \multirow{6}{*}{50/class} &
    65.18\scriptsize{$\pm$0.40} & 
    23.31\scriptsize{$\pm$0.89}  
    \\
BiC~\cite{wu2019bic} && 
    64.63\scriptsize{$\pm$1.27} & 
    22.25\scriptsize{$\pm$1.73}  
    \\
GDumb~\cite{prabhu2020gdumb} && 
    65.90\scriptsize{$\pm$0.28} & 
    - \\
DER++~\cite{PietroBuzzega2020der+} && 
    66.73\scriptsize{$\pm$0.87} & 
    20.67\scriptsize{$\pm$1.24}  
    \\
Co$^2$L~\cite{HyuntakCha2021co2l} && 
    65.90\scriptsize{$\pm$0.14} & 
    23.36\scriptsize{$\pm$0.71}  
    \\ 
\hline
FT-seq & \multirow{5}{*}{0} & 
    28.87\scriptsize{$\pm$1.36} & 
    63.80\scriptsize{$\pm$1.50}  
    \\
EWC~\cite{kirkpatrick2017EWC} && 
    35.00\scriptsize{$\pm$0.43} & 
    56.16\scriptsize{$\pm$0.88}  
    \\
LwF~\cite{li2017LWF} && 
    38.54\scriptsize{$\pm$1.23} &
    52.37\scriptsize{$\pm$0.64}
    \\
L2P~\cite{wang2022L2P} && 
    61.57\scriptsize{$\pm$0.66} & 
    9.73\scriptsize{$\pm$0.47}  
    \\
DualP~\cite{wang2022DualP} && 
    68.13\scriptsize{$\pm$0.49} & 
    4.68\scriptsize{$\pm$0.20} 
    \\ 
\cdashline{1-4}[0.8pt/2pt]
% PT-SNMP &\multirow{4}{*}{0}&
%     49.73\scriptsize{$\pm$0.00} &
%     6.51\scriptsize{$\pm$0.00}
%     \\
% FTF-SNMP &&
%     66.82\scriptsize{$\pm$0.00} &
%     5.46\scriptsize{$\pm$0.00}
%     \\
SDC-SNMP &\multirow{2}{*}{0}&
    67.90\scriptsize{$\pm$0.49} &
    15.44\scriptsize{$\pm$0.84}
    \\
% Baseline4 &&
%     67.30\scriptsize{$\pm$0.00} &
%     7.35\scriptsize{$\pm$0.00}
%     \\
Ours && 
    \bf{72.82\scriptsize{$\pm$0.30}} & 
    \bf{3.90\scriptsize{$\pm$0.21}}  
    \\
\hline
\hline
Upper bound & - & 
    81.95\scriptsize{$\pm$0.11}  & - 
    \\ 
\hline
\end{tabular}
\end{center}
\caption{Result on Split ImageNet-R for CIL. %Memory-based methods get results at buffer sizes of 50/class and 10/class. 0 represents no buffer required.
%\textbf{Bold}: the best data-free result.
}
\label{table:image}
\end{table}

% ################################################
% 【Sequential 5-Datasets】
% ################################################
\begin{table}[]\small
\setlength{\tabcolsep}{5pt}
\begin{center}
\begin{tabular}{lccc}
\hline
Method & Buffer size & FAA ($\uparrow$) & FF ($\downarrow$) \\ 
\hline
ER~\cite{rolnick2019ER} & \multirow{5}{*}{5/class} & 
    80.32\scriptsize{$\pm$0.55} & 
    15.69\scriptsize{$\pm$0.89}  
    \\
BiC~\cite{wu2019bic} && 
    78.74\scriptsize{$\pm$1.41} & 
    21.15\scriptsize{$\pm$1.00}  
    \\ 
DER++~\cite{PietroBuzzega2020der+} && 
    80.81\scriptsize{$\pm$0.07} & 
    14.38\scriptsize{$\pm$0.35}  
    \\ 
Co$^2$L~\cite{HyuntakCha2021co2l} && 
    82.25\scriptsize{$\pm$1.17} & 
    17.52\scriptsize{$\pm$1.35}  
    \\
\hline
ER~\cite{rolnick2019ER} & \multirow{5}{*}{10/class} &
    84.26\scriptsize{$\pm$0.84} & 
    12.85\scriptsize{$\pm$0.62}  
    \\
BiC~\cite{wu2019bic} && 
    85.53\scriptsize{$\pm$2.06} & 
    10.27\scriptsize{$\pm$1.32}  
    \\
DER++~\cite{PietroBuzzega2020der+} && 
    84.88\scriptsize{$\pm$0.57} & 
    10.46\scriptsize{$\pm$1.02}  
    \\
Co$^2$L~\cite{HyuntakCha2021co2l} && 
    86.05\scriptsize{$\pm$1.03} & 
    12.28\scriptsize{$\pm$1.44}  
    \\ 
\hline
FT-seq & \multirow{5}{*}{0} & 
    20.12\scriptsize{$\pm$0.42} & 
    94.63\scriptsize{$\pm$0.68}  
    \\
EWC~\cite{kirkpatrick2017EWC} && 
    50.93\scriptsize{$\pm$0.09} & 
    34.94\scriptsize{$\pm$0.07}  
    \\
LwF~\cite{li2017LWF} && 
    47.91\scriptsize{$\pm$0.33} & 
    38.01\scriptsize{$\pm$0.28}  
    \\
L2P~\cite{wang2022L2P} && 
    81.14\scriptsize{$\pm$0.93} & 
    4.64\scriptsize{$\pm$0.52} 
    \\
DualP~\cite{wang2022DualP} && 
    \bf{88.08\scriptsize{$\pm$0.36}} & 
    \bf{2.21\scriptsize{$\pm$0.69}} 
    \\
\cdashline{1-4}[0.8pt/2pt]
% PT-SNMP &\multirow{4}{*}{0}& 
%     62.24\scriptsize{$\pm$0.00} & 
%     0.18\scriptsize{$\pm$0.00}   
%     \\ 
% FTF-SNMP && 
%     66.63\scriptsize{$\pm$0.00} & 
%     \bf{0.068\scriptsize{$\pm$0.00}}   
%     \\ 
SDC-SNMP &\multirow{2}{*}{0}& 
    80.90\scriptsize{$\pm$0.87} & 
    22.15\scriptsize{$\pm$0.44}   
    \\ 
% Baseline4 && 
%     71.20\scriptsize{$\pm$0.00} & 
%     35.83\scriptsize{$\pm$0.00}   
    % \\ 
Ours && 
    \bf{88.02\scriptsize{$\pm$0.25}} & 
        5.27\scriptsize{$\pm$0.20}   
    \\ 
\hline
\hline
Upper bound & - & 
    95.87\scriptsize{$\pm$0.00} & 
    - \\ 
\hline
\end{tabular}
\end{center}
\caption{Result on Sequential 5-Datasets for CIL. 
%Memory-based methods get results at buffer sizes of 5/class and 10/class, which is enough for this dataset. 0 represents no buffer required.
%\textbf{Bold}: the best data-free result.
}
\label{table:5dataset}
\end{table}

% ################################################
% 【Core50】
% ################################################
\begin{table}[]\small
\setlength{\tabcolsep}{12pt}
\begin{center}
\begin{tabular}{lccc}
\hline
Method & Buffer size & FAA ($\uparrow$) \\ 
\hline
ER~\cite{rolnick2019ER} & \multirow{6}{*}{50/class} & 
    80.10\scriptsize{$\pm$0.56} 
    \\
GDumb~\cite{prabhu2020gdumb} && 
    74.92\scriptsize{$\pm$0.25} 
    \\
BiC~\cite{wu2019bic} && 
    79.28\scriptsize{$\pm$0.30} 
    \\
DER++~\cite{PietroBuzzega2020der+} && 
    79.70\scriptsize{$\pm$0.44} 
    \\
Co$^2$L~\cite{HyuntakCha2021co2l} && 
    79.75\scriptsize{$\pm$0.84}  
    \\
L2P~\cite{wang2022L2P} && 
    81.07\scriptsize{$\pm$0.13} 
    \\ 
\hline
EWC~\cite{kirkpatrick2017EWC} & \multirow{4}{*}{0} & 
    74.82\scriptsize{$\pm$0.60}  
    \\
LwF~\cite{li2017LWF} && 
    75.45\scriptsize{$\pm$0.40}  
    \\
L2P~\cite{wang2022L2P} && 
    78.33\scriptsize{$\pm$0.06}  
    \\
S-Prompts~\cite{wang2022SPrompts} &&
    83.13\scriptsize{$\pm$0.51}  
    \\
\cdashline{1-3}[0.8pt/2pt]
% PT-SNMP & \multirow{4}{*}{0} & 
%     78.61\scriptsize{$\pm$0.00}  
%     \\
% FTF-SNMP && 
%     87.95\scriptsize{$\pm$0.00}  
%     \\
SDC-SNMP && 
    89.98\scriptsize{$\pm$0.74}  
    \\
% Baseline4 && 
%     87.68\scriptsize{$\pm$0.00}  
%     \\
Ours &&
    \bf{92.18\scriptsize{$\pm$0.19}}
    \\
\hline
\hline
Upper bound & - & 
    92.20\scriptsize{$\pm$0.27}
    \\ 
\hline
\end{tabular}
\end{center}
\caption{Result on Sequential Core50 for DIL. %Memory-based methods get results at buffer sizes of 50/class. 0 represents no buffer required.
%\textbf{Bold}: the best data-free result.
}
\label{table:core50}
\end{table}

\begin{table}[] \small
\setlength{\tabcolsep}{2pt}
\begin{center}
\begin{tabular}{ccccc}
\hline
Method  & FAA ($\uparrow$)  & $M$ & Memory(MB) & Inference(ms)\\ 
\hline
L2P (10/class)  & 
    84.21\scriptsize{$\pm$0.53}  & - & 5.01 & 699.5\\
L2P (50/class)  &
    86.31\scriptsize{$\pm$0.59}  & - & 17.3 & 699.5\\
L2P~\cite{wang2022L2P} & 
    83.86\scriptsize{$\pm$0.28}  & - & 1.94 & 699.5\\
DualP~\cite{wang2022DualP} & 
    86.51\scriptsize{$\pm$0.33}  & - & 1.90 & 656.5\\
\hline
\multirow{6}{*}{Ours} & 86.33\scriptsize{$\pm$0.31}  & 1 & 0.31 & 315.9\\
 & 86.36\scriptsize{$\pm$0.24} & 2 & 0.61 & 315.9\\
 & 86.52\scriptsize{$\pm$0.21} & 3 & 0.92 & 316.0\\
 & 87.57\scriptsize{$\pm$0.27} & 4 & 1.23 & 316.1\\
 & 87.52\scriptsize{$\pm$0.30} & 5 & 1.54 & 316.3\\
 & 87.87\scriptsize{$\pm$0.24} & 6 & 1.84 & 316.4\\
\hline
\hline
Upper bound & 91.98\scriptsize{$\pm$0.07} & - & 0 & 315.9\\
\hline
\end{tabular}
\end{center}
\caption{Additional Memory Consumption and Inference Cost. L2P (10/class) and L2P-50 (50/class) represent the results of L2P with buffers of 10/class and 50/class. $M$ is the total number of prototypes per class used in our method. Memory represents the additional memory cost after finishing the last task, compared with the upper-bound model. Inference represents the inference time of different methods (batch size:128, GPU: One Nvidia 3090). 
}
\label{table:memory}
\end{table}

\noindent\textbf{Ablation Study and Hyper-parameter Analysis}. The most important hyper-parameter of our method is the number of saved prototypes for each class. Generally, more prototypes can better preserve the feature distribution of the old classes, but also increases the running memory consumption. Noting that we discard the A-prompts after the prototype update. It thus does not increase the storage memory of the saved model. Tab.~\ref{table:memory} analyzes the additional memory cost of prompt-based methods compared to the non-incremental \emph{upper bound}. We can easily observe that: 1)  Saving prototypes in the feature space is much more memory-efficient than saving the original images; with only 1.8\% (0.31M v.s. 17.3M) memory consumption, we can achieve the same performance as L2P (50/class). 2) Our method is also more memory-efficient than the best data-free method, DualP,  with only a 48.4\% memory needed to achieve the same performance (M=3). 3) Our inference cost is less than half that of DualP and L2P, and almost equal to the joint training upper bound. However, our training time is longer than DualP and L2P, because of the additional analogical making stage. According to our experiments, training the A-prompt for each class takes about 30 seconds on CIFAR-100 (one Nvidia 3090). The training overhead can be further reduced with multiple
GPUs, as the A-prompts are independent of each other. The reduction in inference costs is more economical than the acceptable increase in training costs to implement our method in real applications.

\begin{table}[] \small
\setlength{\tabcolsep}{9pt}
\begin{center}
\begin{tabular}{cccc}
\hline
hyperparameter & value & FAA ($\uparrow$) & FF ($\downarrow$)\\ 
\hline
\multirow{4}{*}{$K$} 
& 10      & 87.69\scriptsize{$\pm$0.17} & 3.08\scriptsize{$\pm$0.24} \\
& 20      & 87.70\scriptsize{$\pm$0.15} & 3.02\scriptsize{$\pm$0.25} \\
& \bf{50}      & 87.87\scriptsize{$\pm$0.24} & 2.78\scriptsize{$\pm$0.07}  \\
& 100     & 87.84\scriptsize{$\pm$0.23} & 2.32\scriptsize{$\pm$0.12} \\ 
\hline
\multirow{4}{*}{$J$} 
& 1  & 87.06\scriptsize{$\pm$0.29} & 3.42\scriptsize{$\pm$0.30} \\
& \bf{5}  & 87.87\scriptsize{$\pm$0.24} & 2.78\scriptsize{$\pm$0.07}  \\
& 10  & 87.88\scriptsize{$\pm$0.29} & 2.69\scriptsize{$\pm$0.23} \\
& 15  & 87.79\scriptsize{$\pm$0.20} & 2.21\scriptsize{$\pm$0.09} \\ 
\hline
\end{tabular}
\end{center}
\caption{Hyperparameter study on Split CIFAR-100. \textbf{Bold}: the selected value in other experiments. }
\label{table:hyper}
\end{table}

\begin{table}[] \small
\setlength{\tabcolsep}{6pt}
\begin{center}
\begin{tabular}{cccc}
\hline
Distance & Scaling Factor & FAA ($\uparrow$) & FF ($\downarrow$)\\ 
\hline
Euclidean 
& -  & 86.89\scriptsize{$\pm$0.29} & 4.13\scriptsize{$\pm$0.46} \\
\cdashline{1-4}[0.8pt/2pt]
\multirow{4}{*}{SN Euclidean} 
& 10  & 87.03\scriptsize{$\pm$0.37} & 3.63\scriptsize{$\pm$0.25} \\
& \bf{20}  & 87.87\scriptsize{$\pm$0.24} & 2.78\scriptsize{$\pm$0.07}  \\
& 30  & 87.46\scriptsize{$\pm$0.38} & 3.23\scriptsize{$\pm$0.19} \\
& 40  & 87.12\scriptsize{$\pm$0.27} & 3.57\scriptsize{$\pm$0.17} \\ 
\hline
\end{tabular}
\end{center}
\caption{Distance metric study on Split CIFAR-100. \textbf{Bold}: the selected value in other experiments, SN Euclidean represents the scaling normalized euclidean distance.}
\label{table:dis}
\end{table}

We also study other hyperparameters used in our method. As shown in Tab.~\ref{table:hyper}, Our method performs consistently over a wide range of values. Therefore, we just select the medium values that perform relatively well in other experiments. Tab.~\ref{table:dis} shows that the scaling normalized euclidean distance is better than the original euclidean distance. We list the ablation study results on the prompt training stage and the model finetuning stage in Tab.~\ref{table:ablation}. As can be seen, $\mathcal{L}_{PP}$ contributes more to the final performance than $\mathcal{L}_{DE}$; $\mathcal{L}_{SC}$ improves FAA by $4.08\%$.

\begin{table}[!t] \small
\setlength{\tabcolsep}{8pt}
\begin{center}
\begin{tabular}{ccccc}
\hline
\multicolumn{5}{c}{The Prompt Training Stage ($\mathcal{L}_{PT}$)} \\
$\mathcal{L_{CC}}$ & $\mathcal{L_{PP}}$ & $\mathcal{L}_{DE}$ & FAA ($\uparrow$) & FF ($\downarrow$)\\ 
\checkmark &&& 82.16\scriptsize{$\pm$0.77} & 12.74\scriptsize{$\pm$0.53} \\
\checkmark & \checkmark && 87.00\scriptsize{$\pm$0.36} &  3.05\scriptsize{$\pm$0.21}\\
\checkmark & \checkmark & \checkmark &  \bf{87.87\scriptsize{$\pm$0.24}} & 
    \bf{2.78\scriptsize{$\pm$0.07}}  \\
\hline
\hline
\multicolumn{5}{c}{The Model Finetuning Stage ($\mathcal{L}_{FT})$} \\
$\mathcal{L}_{LS}$ & $\mathcal{L}_{SC}$ && FAA ($\uparrow$) & FF ($\downarrow$)\\ 
\checkmark &&& 83.79\scriptsize{$\pm$0.35} &  8.74\scriptsize{$\pm$0.22} \\
\checkmark & \checkmark && \bf{87.87\scriptsize{$\pm$0.24}} & 
    \bf{2.78\scriptsize{$\pm$0.07}}   \\
\hline
\end{tabular}
\end{center}
\caption{Ablation study on the loss function.}
\label{table:ablation}
% \vspace{-0.5cm}
\end{table}

\section{Conclusion and Limitation}
%\noindent \textbf{Conclusion}. 
This paper proposes a novel data-free incremental learning method inspired by human analogy capabilities. Instead of saving the original old data, we adopt analogical prompts to reminisce about the previously seen classes from the new-task data. On this basis, we can estimate and counteract the representation shift for the old prototypes, which is critical for unbiased classification. Our method achieves state-of-the-art performance on four incremental learning benchmarks, while having a much lower additional storage requirement than data-replay methods.

\noindent \textbf{Limitation}. The performance of our method is weakened when the semantic gap between tasks is too large. \XP{More efforts shall be made before our method can be } applied to scenarios where there are no task boundaries in the training stage, such as task-free continual learning~\cite{aljundi2019task,wang2022improving}, blurred  boundary  continual  learning~\cite{bang2021rainbow,PietroBuzzega2020der+}, and online learning~\cite{aljundi2019gradient}. 

%Our method cannot be directly
%PietroBuzzega2020der

{\small
\bibliographystyle{ieee_fullname}
\bibliography{egbib}
}

\end{document}